\documentclass[10pt,twocolumn,letterpaper]{article}

\usepackage{wacv}
\usepackage{times}
\usepackage{epsfig}
\usepackage{graphicx}
\usepackage{amsmath}
\usepackage{amssymb}
\usepackage{bm}

\usepackage{caption}
\usepackage{subcaption}
\usepackage{pgfplots}
\pgfplotsset{compat=1.14}
\usepackage{comment}
\usepackage{array, makecell} 
\usepackage[numbers,sort,compress]{natbib}
\def\wacvPaperID{0013} 

\wacvfinalcopy 

\ifwacvfinal
\fi

\ifwacvfinal
\usepackage[breaklinks=true,bookmarks=false]{hyperref}
\else
\usepackage[pagebackref=true,breaklinks=true,colorlinks,bookmarks=false]{hyperref}
\fi

\usepackage{amsmath,amsfonts,bm}

\def\eqref#1{equation~\ref{#1}}

\def\1{\bm{1}}

\DeclareMathAlphabet{\mathsfit}{\encodingdefault}{\sfdefault}{m}{sl}
\SetMathAlphabet{\mathsfit}{bold}{\encodingdefault}{\sfdefault}{bx}{n}

\DeclareMathOperator*{\argmin}{arg\,min}

\begin{document}
\setlength{\abovedisplayskip}{2pt}
\setlength{\belowdisplayskip}{2pt}
\title{Learning Data Augmentation with Online Bilevel Optimization for Image Classification}


\author{Saypraseuth Mounsaveng$^{*1}$, Issam Laradji$^2$, Ismail Ben Ayed$^1$, David V\'azquez$^2$, Marco Pedersoli$^1$\\
$^1$\small{\textit{\'ETS Montreal}}
$^2$\small{\textit{Element AI}}\\
} 

\maketitle

\renewcommand{\thefootnote}{\fnsymbol{footnote}}
\footnotetext[1]{Corresponding author: saypraseuth.mounsaveng.1@etsmtl.net}
\footnotetext[2]{Code is available at \url{https://github.com/ElementAI/bilevel_augment}}

\begin{abstract}
Data augmentation is a key practice in machine learning for improving generalization performance. However, finding the best data augmentation hyperparameters requires domain knowledge or a computationally demanding search. We address this issue by proposing an efficient approach to automatically train a network that learns an effective distribution of transformations to improve its generalization. Using bilevel optimization, we directly optimize the data augmentation parameters using a validation set. This framework can be used as a general solution to learn the optimal data augmentation jointly with an end task model like a classifier. Results show that our joint training method produces an image classification accuracy that is comparable to or better than carefully hand-crafted data augmentation. Yet, it does not need an expensive external validation loop on the data augmentation hyperparameters. 
\end{abstract}

\vspace{-6mm}
\section{Introduction}
\label{sec:Introduction}
Deep learning methods are based on large models in which the number of parameters is much higher than the dimensionality of the input data as well as the number of available samples~\cite{Simonyan2014VGG, He2015ResNet}. In this setting, overfitting is a major problem~\cite{Srivastava2014DropoutAS}. Standard regularization techniques applied directly to the model parameters only add very general knowledge about the parameter values, which leads to modest improvement in the final model accuracy~\cite{Nowlan1992SimplifyingNN}.
Adding training samples artificially generated by applying predefined transformations to the initial samples, which is referred to as data augmentation, has shown to be a promising regularization technique to increase a model performance~\cite{hernandez2018data}. 
However, the selection of the best data augmentation is challenging and requires specific domain knowledge.
\begin{figure}[t]
  \centering
  \includegraphics[width=\linewidth,trim={0.6cm 0 0.6cm 0},clip]{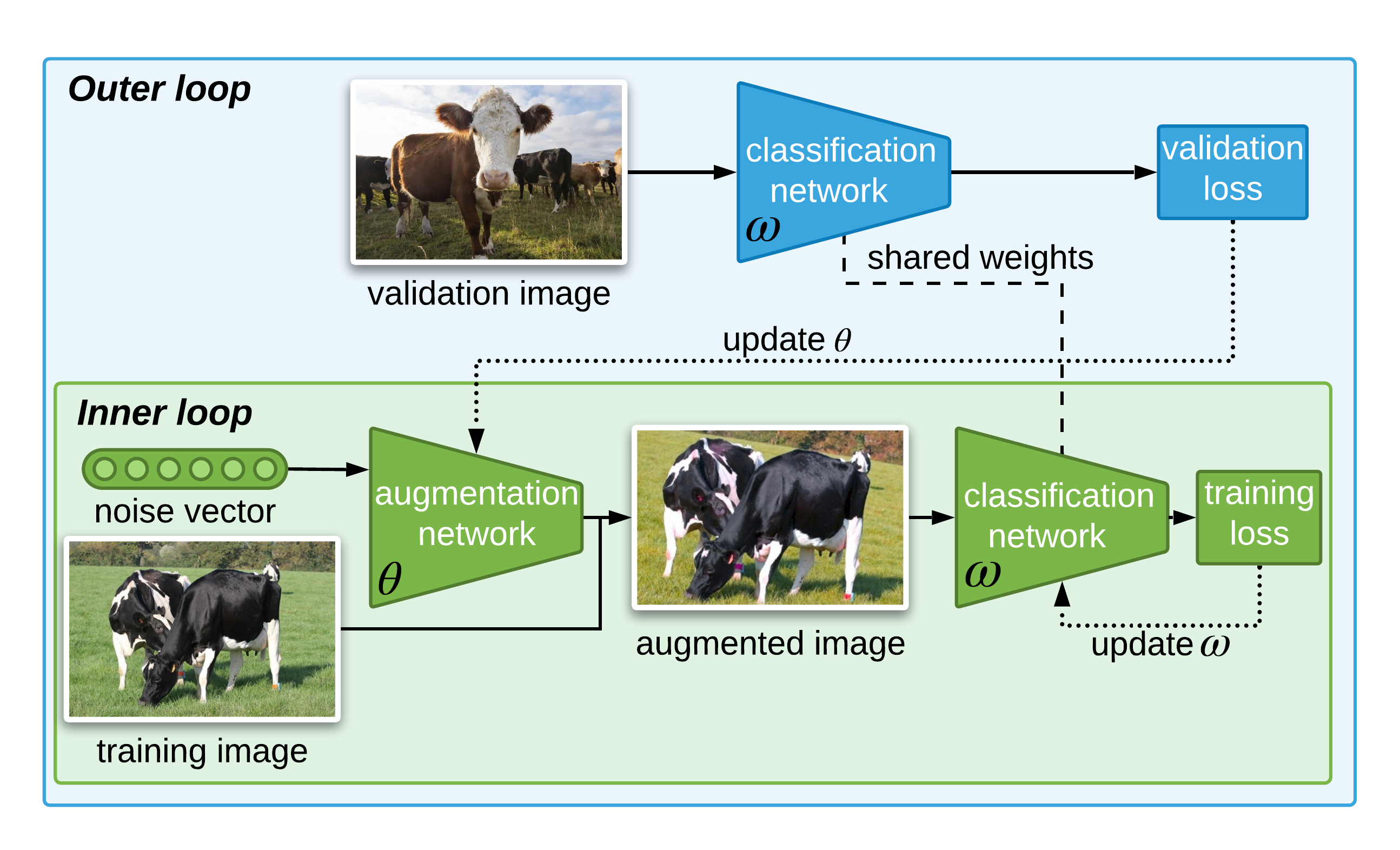}
  \caption{\textbf{Model training.} In an epoch, the classifier parameters $\omega$ are trained in the standard supervised way in the inner loop. Jointly, the data augmentation parameters are trained on the validation set in the outer loop using an online differentiable method.}
  \label{fig:model}
  \vspace{-3mm}
\end{figure}
Indeed, data transformations can be valid only for specific domains and heuristically chosen transformations, for example by transferring transformations useful in a domain into another can be counterproductive. For instance, a data augmentation transformation like horizontal flip is valid for natural images because, in nature, the horizontal mirror of an object is visually still a valid object. However, applied to a dataset containing numbers or letters, it can generate a non-existing symbol or even a different symbol of the alphabet, which would confuse the model training.

A simple way to define the best data augmentation is to use expert knowledge to define the best transformations and their parameters for a given dataset. However, this is not practical, as for each single dataset, an expert should be consulted to obtain a possibly useful set of transformations and their parameters, which is not always possible due to cost constraints or limited expert knowledge availability. 

To mitigate this challenge, it is possible to select those transformations heuristically, and then find their optimal parameters using validation data. Conventionally, a hyperparameter search is performed across different sets of transformations, and the one that leads to the best validation accuracy is selected as the optimal set of transformations. This approach is appealing as it allows the model to learn the best transformations directly from the data, but it is not scalable. Given a large range of transformations to test, retraining the algorithm every time with a different transformation set is computationally very demanding~\cite{Cubuk_2019_CVPR}.

In this work, we address the problem of how to learn the data augmentation that maximizes the validation accuracy efficiently by proposing a method based on bilevel optimization. With this framework, we aim at identifying image transformations that minimize the validation loss while training the end task model. However, as described in section~\ref{sec:learning} and shown in Fig.~\ref{fig:model}, instead of solving the complete bilevel optimization problem, we approximate it with an online version where in every iteration a new set of transformations is learned and adapted to the learning phase.

We summarize our contributions as follows:
i) we propose an online, differentiable approach for learning the optimal data augmentation regime using a validation set. As this method is differentiable, we can efficiently optimize a large transformation network that learns to perform data augmentation automatically;
ii) we show that our proposed model using different sets of transformations achieve comparable or better results than conventional methods on five different datasets.  Further improvements were shown with our method on a medical imaging dataset where effective transformations are difficult to define.

The remainder of the paper is structured as follows.  We conduct a literature review in Section \ref{sec:related_work}. In Section \ref{sec:learning} we explain how to approximate our bilevel optimization problem such that we can jointly learn the optimal data augmentation and the classifier. In Section~\ref{sec:experimental_setup} we define the experimental setup. Finally, we present our experimental results and draw conclusions about the presented work in Section \ref{sec:results}, and \ref{sec:conclusion}, respectively.

\vspace{-2mm}
\section{Related Work}
\label{sec:related_work}
Data augmentation consists in creating new data points from existing ones in order to get a larger training set. It was found to be essential for achieving state-of-the-art image classification results~\cite{hernandez2018deep}.

Data augmentation transformations are usually chosen heuristically based on expert domain knowledge. For natural images, usual transformations are image flip, rotation and color changes ~\cite{perez2017effectiveness}. More complex transformations such as occluding parts of an image~\cite{devries2017improved} or blending images~\cite{zhang2018mixup, lemley2017smart} seem also to be useful. These transformations can significantly improve the task performance. However, there is no guarantee that they are optimal nor that they are even useful at all. 
To avoid the manual selection of transformations, recent studies have investigated automatic data augmentation learning. We distinguish those methods between GAN-based and AutoAugment-based approaches.

\vspace{-3mm}
\paragraph{GAN-based}
Generative Adversarial Networks (GANs)~\cite{Goodfellow2014GenerativeAN} can generate realistic new samples of a certain dataset or class, thus they can be adapted for data augmentation.
\citet{mirza2014conditional} and \citet{DBLP:conf/icml/OdenaOS17} proposed to generate images conditioned on their class that could be directly used to augment a dataset. CatGAN~\cite{springenberg2016iclr} on the other hand, performs unsupervised and semi-supervised learning as a regularized information maximization problem~\cite{krause2010discriminative} with a regularization based on the generated samples. Also based on GAN, but directly used for data augmentation, DAGAN~\cite{antoniou2018augmenting} conditions the augmented image on the input image.
TripleGAN~\cite{chongxuan2017triple} and Bayesian data augmentation~\cite{tran2017bayesian} train a classifier jointly with the generator.
These approaches generate general image transformations, but in practice, it is not as performant as using predefined transformations.
TANDA~\cite{ratner2017learning} is the only GAN-based approach that uses predefined transformations. It defines a large set of transformations and learns how to combine them to generate new samples that follow the same distribution as the original data. This approach is better, but it is still based on the assumption that the augmented data should follow the same distribution as the original data. Instead, we argue that data augmentation should improve the performance of the classifier, independently from the visual similarity of the generated data.

\vspace{-3mm}
\paragraph{AutoAugment}
AutoAugment~\cite{Cubuk_2019_CVPR} is a data augmentation method that learns sequences of transformations that maximize the classifier accuracy on a validation set. This objective is better than simply reproducing the same data distribution as in GAN-based models, as it favors transformations that generalize well on unseen data. 
However, it is computationally expensive as it performs the complete bilevel optimization by training the classifier in the inner loop until convergence for each set of evaluated transformations. 
Some solutions to reduce the computational cost are proposed in follow-up works. Fast AutoAugment~\cite{lim2019fast} optimizes the search space by matching the density between the training set and the augmented data. Alternatively, Population Based Augmentation (PBA)~\cite{ho2019pba} focuses on learning the optimal augmentation schedule rather than only the transformations. However, even if these approaches reduce the computational cost of AutoAugment, they do not leverage gradient information. Faster AutoAugment~\cite{Hataya2019FasterAL} does this by combining AutoAugment with a GAN discriminator and considering transformations as differentiable functions. OHL-Auto-Aug \cite{Lin2019OnlineHL} uses an online bilevel optimization approach and the REINFORCE algorithm on an ensemble of classifiers to estimate the gradient of the validation loss and learn an augmentation probability distribution.
RandAugment~\cite{Cubuk2019RandAugmentPD} goes further by showing that a same performance level as AutoAugment can be obtained by randomly selecting transformations from the predefined pool and just tune the number of transformations to use and a global (same for all transformations) magnitude factor. However, this approach also requires prior knowledge of useful transformations.
\vspace{-3mm}
\paragraph{Hyperarameter Learning}
Our work has similarities with techniques used in the hyperparameter optimization field. 
Hyperparameters tuning is important to obtain the best performances when training neural networks on a given dataset. Classic approaches assume that the learning model is a black-box and use methods like grid search, random search~\cite{NIPS2011_4443, bergstra2013making}, Bayesian optimisation~\cite{snoek12practical}, or a tree-search approach~\cite{hutter2011sequential}. These approaches are simple but expensive because they repeat the optimization from scratch for each sampled value of the hyperparameters and so are only applicable to low dimensional hyper-parameter spaces. 
A different line of research is to leverage the gradient of these (continuous) hyperparameters (or hyper-gradients) to perform the hyper-optimization.
The first work proposing this idea~\cite{bengio2000gradient}, shows that the implicit function theorem can be used to this aim.
\cite{domke2012generic} was the first work to propose a gradient-based method using a bilevel optimization approach~\cite{colson2007overview} to learn hyperparameters. Using a bilevel optimization approach to train a neural network is challenging, as usually there is no closed-form expression of the function learned in the inner loop (Section~\ref{sec:learning}).
To address this, \citet{pmlr-v37-maclaurin15} and later \citet{pmlr-v70-franceschi17a} proposed methods to reverse the forward pass to compute the gradient of the validation loss. However, these methods are applicable only when the number of hyperparameters and the complexity of the models are limited due to the memory needed to save the intermediate steps.
Another approach to address the computational hurdle in the inner loop is to calculate an approximation of the gradient like in \citet{pmlr-v48-pedregosa16} \citet{luketina2016scalable} or \citet{mackay2019self}. Our method differentiates from those by using truncated back propagation to estimate the gradient of the validation loss.
Finally, note that hyper-parameter optimization presents some similarities to meta learning as shown in \citet{pmlr-v80-franceschi18a}. For instance, in MAML~\cite{finn2017model}, a shared model initialization is learned to minimize the validation loss and therefore improve the generalization capabilities of the model.


\begin{figure*}[t]
\captionsetup[subfigure]{aboveskip=5pt,belowskip=-3pt}
\centering
\begin{subfigure}[t]{0.55\textwidth}
\includegraphics[height=7cm,trim={0 4cm 0 4cm}]{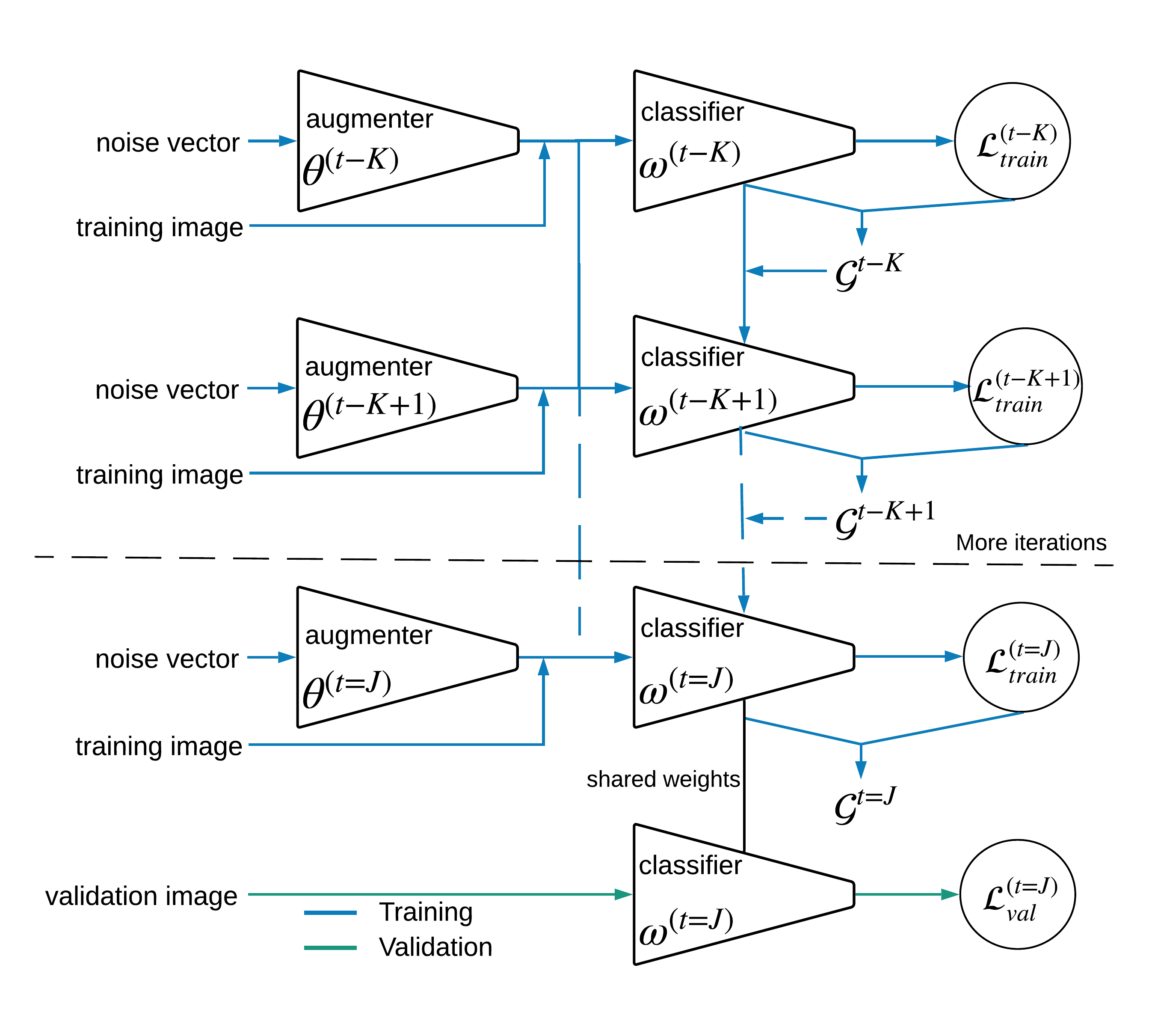}
\caption{Forward pass.}\label{fig:forward}
\end{subfigure}
\hspace*{\fill}
\begin{subfigure}[t]{0.44\textwidth}
\includegraphics[height=7cm,trim={0 6cm 0 5cm}]{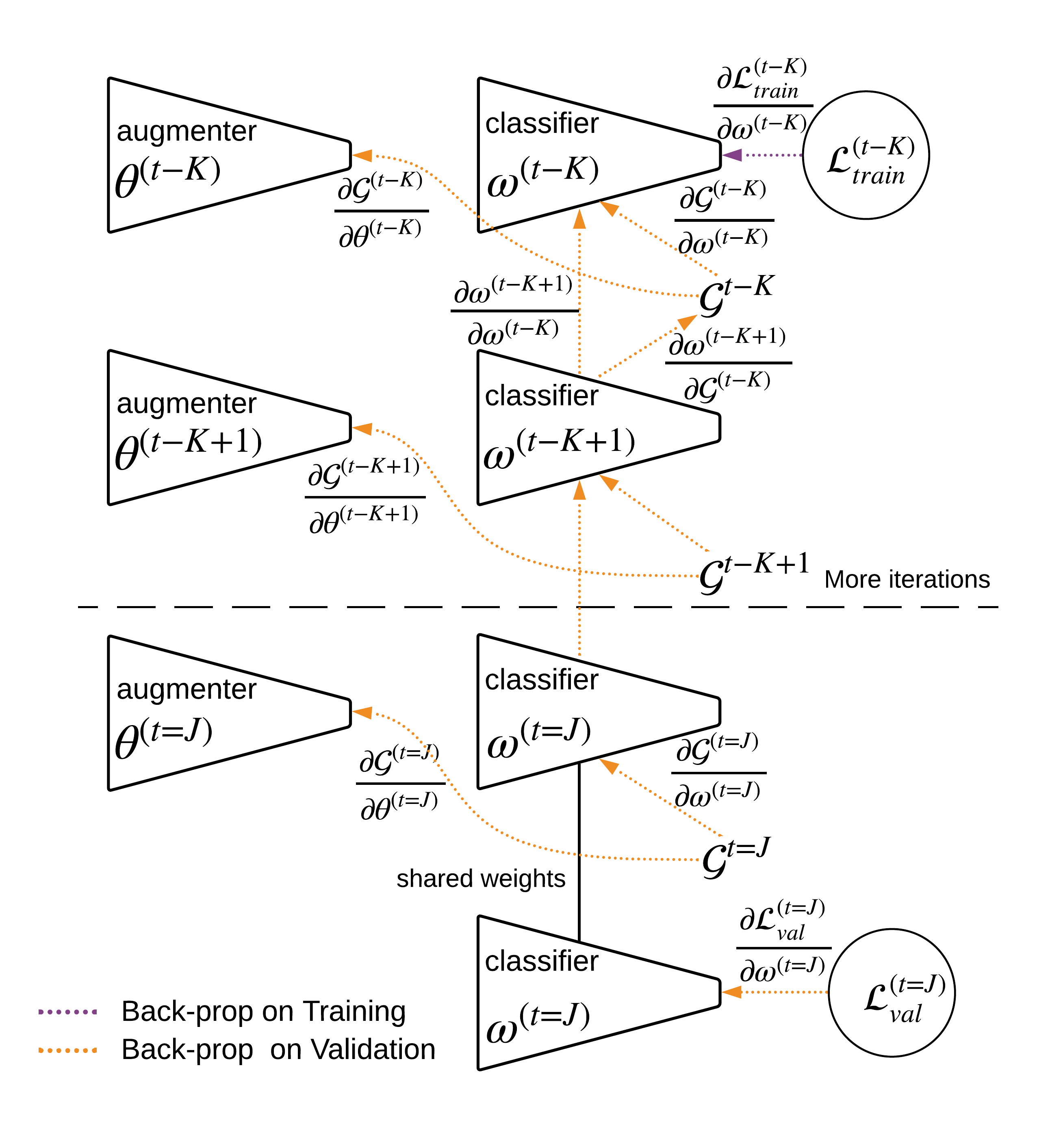}
\caption{Backward pass.}\label{fig:backward}
\end{subfigure}
\caption{\textbf{Computational graph of our model at iteration $\boldsymbol{t=J}$.}
$K$ is the number of gradient unfolding steps, and J is the number of inner loop iterations after which $\theta$ gets updated. The case where K=J=T (T being the iteration of the classifier convergence) is the complete bilevel optimization as in Eq.\ref{equ:bilevel1} whereas K=J=1 corresponds to updating $\theta$ at each mini-batch ($K=1)$, suing only one step of gradient unfolding ($J=1$).}
\label{fig:graph}
\end{figure*}

\vspace{-2mm}
\section{Proposed Data Augmentation Method}
\label{sec:learning}
Consider a labeled set $\mathcal{X}:=\{x_i,y_i\}^{N}_{i=1}$, where $x_i$ is an input image, $y_i$ the associated class label, N the number of samples and $\hat{\mathcal{X}}$ the set of transformed images. We formulate the problem of identifying effective data augmentation transformations as a bilevel optimization problem. In this setup, the augmenter $\mathcal{A}_\theta:\mathcal{X}\rightarrow{\hat{\mathcal{X}}}$ is parametrized by ${\theta}$ and is used to minimize the loss $\mathcal{L}$ on the validation data $\mathcal{X}_{val}$ in the outer loop. In the inner loop, the classifier parameters $\omega$ are optimized  on the training data $\mathcal{X}_{tr}$ in the standard supervised way. This formulation can be written as:
\begin{align}
\theta^* &= \argmin_\theta
\mathcal{L}(\mathcal{X}_{val}, \omega^*) 
\label{equ:bilevel1} \\
s.t. \quad \omega^* &= \argmin_{\omega} \mathcal{L}(\mathcal{A}_{\theta}(\mathcal{X}_{tr}),\omega).
\label{equ:bilevel2}
\end{align}
While optimizing a few hyperparameters on the validation data is feasible with black-box approaches such as grid and random search~\cite{bergstra12random} or Bayesian optimization~\cite{snoek12practical}, it is not efficient. With bilevel optimization, our aim is to efficiently learn an entire neural network $\mathcal{A}_\theta$ (possibly with thousands of parameters $\theta$) which defines a distribution of transformations that should be applied on the training data to improve generalization. 

Gradient descent was shown to be an efficient method for optimizing parameters of large networks. In problems such as architecture search~\cite{liu2018darts}, the parameters can be directly optimized with gradient descent (or second order methods) against the training and validation data.
However, this is not the case for data augmentation. The reason is that the transformation network $\mathcal{A}_{\theta}$ is optimized to maximize the validation score, but applies transformations only on the training set. Therefore, first order methods would not work. The aim of data augmentation is to introduce transformations during the training phase that can make the model invariant or partially invariant to any transformations that can occur at test time. If we optimize the transformation network directly on the validation data, the model will simply select trivial solutions such as the identity transformation. This approach has been used for  object localization~\cite{jaderberg2015spatial} and it did not improve the model generalization performance as much as data augmentation.
To solve this issue, new methods relied on reinforcement learning instead of gradient descent to learn effective data augmentation~\cite{Cubuk_2019_CVPR, lim2019fast, ho2019pba}.

In this work, we show that in the case of a differentiable augmenter $\mathcal{A}_\theta$, there is a simple, efficient way to find optimal data transformations based on gradient descent that generalize well on validation data. We formulate our problem as an approximation to bilevel optimization by using truncated back-propagation as it allows our method to:
i) efficiently estimate a large number of parameters to generate the optimal data augmentation transformations by gradient descent;
ii) obtain an online estimation of the optimal data augmentation during the different phases of the training, which can also be beneficial~\cite{Golatkar2019TimeMI};
iii) change the training data to adapt to different validation conditions as in supervised domain adaptation.\\
Although approximate bilevel optimization has already been proposed for hyperparameter optimization~\cite{Shaban-AISTATS-19, pmlr-v80-franceschi18a, pmlr-v70-franceschi17a}, in this paper we show that it can be used for training a large, complex model (the augmenter $\mathcal{A}_\theta$ network) in order to learn an effective distribution of transformations.

\vspace{-3mm}
\subsection{Approximate Online Bilevel Optimization}
As shown in Eq.~\ref{equ:bilevel1} and \ref{equ:bilevel2}, the problem of finding the optimal data augmentation transformations $\mathcal{A}_\theta$ can be cast as a bilevel optimization problem. This problem can be solved by iteratively solving Eq.~\ref{equ:bilevel2} to find the optimal network weight $\omega^*$, given the parameters of the transformation $\theta$ and then updating $\theta$:
\begin{equation}
\footnotesize
\theta \leftarrow \theta - \eta_\theta\nabla_\theta \mathcal{L}(\mathcal{X}_{val},\omega^*)
\end{equation}
where $\eta_\theta$ is the learning rate used to train the augmenter network.\\ 
However, as the augmentations are to be applied only on the training dataset and not on the validation set, calculating $\frac{\partial{\mathcal{L}(\mathcal{X}_{val},\omega^*)} }{\partial \theta}$ is not trivial. To enable this calculation, we use the fact that the weights $\omega$ of the network are shared between training and validation data and use the chain rule to differentiate the validation loss $\mathcal{L}(\mathcal{X}_{val},\omega^*)$ with respect to the hyperparameters $\theta$. In other words, instead of using a very slow black-box optimization for $\theta$, we can exploit gradient information 
because the model parameters $\omega^*$ are shared between the validation and the training loss.\\
We define the gradient of the validation loss with respect to $\theta$ as follows:
\begin{equation} \label{equ:grad1}
\footnotesize
\begin{split}
{\nabla_\theta \mathcal{L}}(\mathcal{X}_{val},\omega^*)&=\frac{\partial{\mathcal{L}(\mathcal{X}_{val},\omega^*)} }{\partial \theta}\\
&= \frac{\partial{\mathcal{L}}(\mathcal{X}_{val},\omega^*)}{\partial \omega^*} \frac{\partial{\omega^*}}{\partial \theta}
\end{split}
\end{equation}
By defining $\mathcal{G}^{(t)}$ as the gradient of the training loss at iteration $t$:
\begin{equation}
\footnotesize
\mathcal{G}^{(t)}=\nabla_{\omega}\mathcal{L}(\mathcal{A}_{\theta}(\mathcal{X}_{tr}),\omega^t)
\end{equation}
we can write $\frac{\partial{\omega^*}}{{\partial \theta}}$ in Eq.~\ref{equ:grad1} as:
\begin{equation}
\footnotesize
\frac{\partial{\omega^*}}{\partial \theta} = \sum_{i=1}^{T-1} \frac{\partial{\omega^{(T)}}}{\partial \omega^{(i)}} \frac{\partial{\omega^{(i)}}}{\partial{\mathcal{G}^{(i-1)}}}\frac{\partial{\mathcal{G}^{(i-1)}}}{\partial {\theta}}
\end{equation}
where T is the iteration when the classifier converges.

As $\omega^*$ represents the model weights at training convergence, they depend on $\theta$ for each iteration of gradient descent. Thus, to compute $\frac{\partial{\omega^*}}{\partial \theta}$, one has to back-propagate throughout the entire $T$ iterations of the training cycle. An example of this approach is in \citet{pmlr-v37-maclaurin15}.
This approach is feasible only for small problems due to the large requirements in terms of computation and memory. 
However, as optimizing $\omega^*$ is an iterative process, instead of computing $\frac{\partial{\omega}}{\partial \theta}$ only at the end of the training loop, we can estimate it at every iteration $t$: 
\begin{equation}
\footnotesize
\frac{\partial{\omega^*} } {\partial \theta} \approx
\frac{\partial{\omega^{(t)} } } {\partial \theta^{(t)}} = \sum_{i=1}^{t} \frac{\partial{\omega^{(t)}}}{\partial{\omega^{(i)}}} \frac{\partial{\omega^{(i)} }} {\partial{\mathcal{G}^{(i-1)}}}\frac{\partial{\mathcal{G}^{(i-1)}}}{\partial {\theta^{(i)}}},
\label{equ:online}
\end{equation}
This procedure corresponds to dynamically changing $\theta$ during the training iterations (thus it becomes $\theta^{(t)}$) to minimize the current validation loss based on the training history. Although this formulation is different from the original objective function, adapting the data augmentation transformations dynamically with the evolution of the training process can improve generalization performance~\cite{Golatkar2019TimeMI}.
This relaxation is often used in constrained optimization for deep models, in which constraints are reformulated as penalties and their gradients are updated online, without waiting for convergence, to save computation \cite{Pathak2015ConstrainedCN}. However, in our case, we cannot write the bilevel optimization as a single unconstrained formulation in which the constraint in $\omega^*$ is summed with a multiplicative factor that is maximized (i.e., Lagrange multipliers), because the upper level optimization should be performed only on $\theta$, while the lower level optimization should be performed only on $\omega$. Nonetheless, even with this relaxation, estimating $\frac{\partial{\omega^*} } {\partial \theta}$ still remains a challenge as it does not scale well. Indeed, the computational cost of computing $\frac{\partial{\omega^{(t)}}}{\partial{\theta^{(t)}}}$ grows with the number of iterations $t$ as shown in Eq.~\ref{equ:online}.
To make the gradient computation constant at each iteration we use truncated back-propagation similarly to what is commonly used in recurrent neural networks~\cite{Williams90anefficient}:
\begin{equation} \label{equ:truncated}
\footnotesize
\frac{\partial{\omega^{(t)} } } {\partial \hat\theta} \approx \sum_{i=t-K}^{t} \frac{\partial{\omega^{(t)} } } {\partial{\omega^{(i)} } } \frac{\partial{\omega^{(i)} }} {\partial{\mathcal{G}^{(i-1)}}}\frac{\partial{\mathcal{G}^{(i-1)}}}{\partial {\theta^{(i)}}},
\vspace{-3mm}
\end{equation}
where $K$ represents the number of gradient unfolding that we use. 
Fig.~\ref{fig:graph}b. shows the computational graph used for this computation.\\
Additionally, as \citet{Williams90anefficient}, we consider a second parameter $J$ which defines the number of inner loop training iterations after which $\theta$ is updated, in other words how often the computation of the gradients of $\theta$ is performed. The situation where $K=J=T$ is the exact bilevel optimization as shown in Eq.~\ref{equ:bilevel1} while $K=J=1$ corresponds to updating $\theta$ at each iteration, in our case mini-batch ($K=1$), using only one step of gradient unfolding ($J=1$). A theoretical analysis of the convergence of this approach is presented in \citet{Shaban-AISTATS-19}.

\vspace{-1mm}
\subsection{Augmenter Networks}
\label{subsec:augmenter_networks}
We present augmenter networks that can perform two types of transformations: geometrical and color.\\
\textbf{Geometrical} transformation is a good form of data augmentation, because it simulates the fact that in the real world the same object can be located at multiple positions and seen from different viewpoints.
We use the transformation model of spatial transformer networks~\cite{jaderberg2015spatial}, but for data augmentation instead of data alignment. Thus, as illustrated in Fig.~\ref{fig:model}, the augmenter is composed of a module that generates a set of transformation parameters followed by a module that applies the generated transformation to the original image. 
In our experiments, we consider scenarios where the augmenter network learn affine transformations as well as scenarios where it learns only translation. In this case, only two values are learned (translation values respectively on x and y axis). Note that the learned transformations are not conditioned on the input image but defined only based on random noise.\\
\textbf{Color} transformations considered are: hue, saturation, contrast and brightness. In this case, the augmenter receives as input a random noise vector and generate a single value representing the amplitude of each color transformation. 

\vspace{-2mm}
\section{Experimental Setup}
\label{sec:experimental_setup}
\subsection{Datasets}
In our experiments, we consider the five following datasets:\\
CIFAR10~\cite{krizhevsky2009learning} is a dataset composed of 60,000 32x32 natural color images distributed in 10 different classes (6,000 images per class). This dataset is split into a training set of 50,000 images and a test set of 10,000 images.\\
CIFAR100~\cite{krizhevsky2009learning} is an extension of CIFAR10 dataset. It contains the same number of images at the same resolution, but they are distributed in 100 classes instead of 10.\\
ImageNet~\cite{ILSVRC15} is a dataset of 1.28 million natural color images in the training set and 50,000 images in the test set. The image size is variable, so in our experiments, we resize them to a resolution of 224x224.\\
Tiny ImageNet is a subset of ImageNet~\cite{ILSVRC15} containing 200 classes and images resized to 64x64. Each class has 500 training images, 50 validation images, and 50 test images. Since the test labels are not available, the validation set is used as test set and 20\% of the training set is used for validation.\\
Finally, BACH~\cite{Aresta2019BACHGC} is a dataset of 400 breast cancer histology images of resolution 2048 x 1536 distributed in 4 balanced classes of 100 images. As there is no test set publicly available, we use in our experiments 50\% of the dataset for training and validation and 50\% for test.

\vspace{-1mm}
\subsection{Implementation Details}
Our model is composed of a classifier and an augmenter network.
To facilitate fair comparison of the results, we use in our experiments the same classifiers as in previous works: BadGAN~\cite{dai2017good}, ResNet18~\cite{He2015ResNet}, ResNet50~\cite{He2015ResNet} and Wide-ResNet-28-10~\cite{Zagoruyko2016WideRN}.
BadGAN is a simple CNN based architecture composed of 9 convolutional layers with Leaky ReLUs and a MLP classifier. ResNet18 and ResNet50 are respectively 18 and 50 layers deep neural network with residual connections and WideResNet 28-10 is a ResNet network with 28 layers and a width factor of 10.

The augmenter learning the geometric and color transformations is a MLP that receives a noise vector as input and generates the transformation parameters. We experimented with three sizes.
The \emph{small} network has an input and output size of $n$, $n$ being the number of hyperparameters to optimize (6 for affine transformations, 2 for translation and 1 for each color transformations), and it has two layers with respectively $n$ and $10n$ neurons. The \emph{medium} one has an input size of 100 and two layers of 64 and 32 neurons. The \emph{large} one has an input size of 100 and four layers of 512, 1024, 124 and 512 neurons.
In order to have differentiable affine and color transformations, we use the Kornia~\cite{eriba2019kornia} library and the affine\_grid and grid\_samples functions of the torchvision package of pytorch framework.

In all experiments, we use 20\% of the training set to form the validation set. Although in principle we usually use a separate validation set for training the augmenter, in practice, we noticed that reusing the training data in a variant of this holdout approach (the training set is randomly split into train and validation at each epoch) yields better results. However, it is important that the batch of samples used to learn the augmenter is different from the one used to train the classifier to ensure that the model learns data augmentation parameters that generalize well.
In preliminary experiments, we tried different values for the frequency of updating $\theta$ $J$ and the number of steps of back-propagation $K$, but they did not show relevant improvements. Therefore, for all our experiments, we use $K=J=1$. In practice, the classifier is updated after each training mini-batch and the augmenter after each validation mini-batch. The only predefined transformation used by our model is the horizontal flip (vertical for BACH) as it is not differentiable. 

\vspace{-2mm}
\section{Results}
\label{sec:results}
The goal of our method is to learn data augmentation automatically. Our experiments compare the performance of different classifiers without data augmentation (baselines), the same classifiers with the best-known hyperparameters for data augmentation (predefined), state of the art methods and our method. We experiment with two groups of transformations: geometric and color transformations.

\vspace{-1mm}
\subsection{Geometric Transformations}
In this section, we evaluate our model by investigating learned geometric transformations.

\begin{table}[t]
\footnotesize
\centering
\begin{tabular}{l|c|c|c|}
    ResNet18 / CIFAR10 & \textbf{Trans.} & \textbf{Affine} & \textbf{Cost}\\
    \hline
    Baseline             & 88.55 & 88.55 & 1\\
    Predefined           & 95.28 & 94.59 & $>60$\\
    Transf. invariant (STN)     & 92.14 & 90.31 & 1.1\\
    Validated magnitude        & 94.58 & 93.43 & 11.5\\\hline   
    Our model + HFlip & \textbf{95.35} &  \textbf{95.16} & 5.3\\
\end{tabular}%
\caption{\textbf{Impact and training cost of different geometric data augmentation strategies on classification accuracy on CIFAR10.} Considering only translation and affine transformations, our approach is faster than methods requiring a validation loop and is more efficient than predefined data augmentation, STN and validated magnitude of predefined data augmentation.}
\label{tab:geometric_resnet18_cifar10}
\end{table}

In a first experiment, we assess the computational efficiency of our method and the utility of the learned geometric transformations for the classification task. In Tab.~\ref{tab:geometric_resnet18_cifar10} we compare the performance of our method on CIFAR10 (with ResNet18) against several methods in terms of accuracy and training cost for translation and affine transformations. 
We define our \emph{baseline} as a training without any data augmentation and we consider its training time cost as $1$. 
\emph{Predefined} represents a classifier trained with the usual standard geometric data augmentation: horizontal flip and random translation between -4 and 4 pixels along x and y axis.
To estimate the training cost of this scenario, we consider the general case where the best data augmentation setting is not known and many different values have to be tested using a grid or random search.
For instance, for the 6 parameters of an affine transformation, and 2 different values to try for each parameter, the number of models to validate is $2^6=64$.
In the third case, the augmenter is trained to be \emph{transformation invariant} similarly to the spatial transformer networks~\cite{jaderberg2015spatial}. The transformations generated by the augmenter 
are applied on training as well as on test and the update of the augmenter parameters $\theta$ is done on the same data as the update of the classifier parameters $\omega$.
This approach has a very low computational cost ($1.1$, just the overhead of applying the augmenter) and its accuracy is better than using no data augmentation, but far from a model trained with a good data augmentation.  
Finally, we consider a \emph{validated magnitude} approach that selects only a single parameter defining the magnitude of the transformation parameters from which an actual transformation is sampled from.
This is similar to the strategy used in RandAugment~\cite{Cubuk2019RandAugmentPD}. This performs surprisingly well but is still inferior to our model and with a higher cost for validating the magnitude of the transformations.     
The last row presents the results of our approach.
For both translation only and affine transformations, we obtain better results than the other approaches. In terms of computational cost, our approach is around $5$ times slower than a basic training without data augmentation. 
However, as our approach learns the data augmentation parameters directly and does not need to loop over possible values, it is already 14x faster than the simple case of predefined data augmentation described above where we consider only 2 possible values for each parameter.

\begin{table}[t]
\footnotesize
\centering

\begin{tabular}{l|c|c|c|c}
     CIFAR-10& \multicolumn{2}{c|}{\textbf{BadGAN}} & \multicolumn{2}{c}{\textbf{ResNet18}} \\
    \textbf{Aug.-Class.} & \textbf{Tr.} &  \textbf{Aff.} & \textbf{Tr.} &  \textbf{Aff.}\\
    \hline
    Small                    & 93.65 & 93.62 & \textbf{95.35} & \textbf{95.16}\\
    Medium                   & \textbf{93.75} & \textbf{93.63} & 95.25 & 95.06\\
    Large                    & 93.65 & 93.39 & 95.00 & 94.83\\
\end{tabular}%
\caption{\textbf{Impact of architecture on classification accuracy.} Increasing the classifier size improves the model performance. Increasing the augmenter size has no significant impact on the final classification accuracy.}
\label{tab:augmenter_sizes}
\end{table}

In a second experiment, we investigate the influence of the augmenter network and the classifier size on the performance of a model trained on CIFAR10.
In Tab.~\ref{tab:augmenter_sizes}, results show that a larger classifier (from BadGAN to ResNet18) improves the performance. However, the size of the augmenter network does not have a significant impact on the accuracy of the classifier. Thus, in the following experiments, we use the \emph{small} augmenter, which is faster to train.

In a third experiment, we investigate the efficiency of learned transformations against heuristically chosen ones. For this, we consider the medical imaging dataset BACH, as possibly useful data augmentation for histological images is not trivial to define as opposed to natural images. The usual heuristically chosen geometric transformations for medical images are vertical flip and affine transformations.
\begin{table}[t]
\footnotesize
\centering
\begin{tabular}{l|c}
    {ResNet18 / BACH} & Acc.\\
    \hline
    Baseline & 49.50\\
    Baseline + only VFlip & 46.00\\
    Baseline + VFlip + Affine & 50.60\\
    \hline
    Our model(affine) + VFlip & \textbf{56.00}\\
\end{tabular}%
\caption{\textbf{Impact of geometric data augmentation on classification accuracy on BACH.} Vertical flip alone decreases the model performance. Used in combination with affine transformations, it improves the classification accuracy. Best performances are obtained using learned affine transformations.}
\label{tab:geometric_bach}
\end{table}
In Tab.~\ref{tab:geometric_bach}, we compare our model learning affine transformations to a \emph{baseline} trained without data augmentation, with \emph{vertical flip only} and finally with \emph{vertical flip and predefined affine transformations}: random translation between -4 and 4 pixels along x and y axis, random scale with factor  between 0.5 and 2 and rotation between -10 and 10 degrees. Results show that only using vertical flip, which is a common transformation in medical imaging is reducing the performance of the classifier whereas using a broader range of affine transformations is yielding a better model performance than using no augmentation. Our model, using only vertical flip as predefined augmentation, obtains a much better final accuracy, which shows that the learned transformations are more useful for the classifier than the hand defined ones.

\vspace{-1mm}
\subsection{Color Transformations}
In this section, we investigate the impact of color transformations alone and in combination with affine transformations on different datasets. 
\begin{table}[t]
\footnotesize
\centering
\begin{tabular}{l|c|c}
    ResNet18 / CIFAR10 &  Acc.\\
    \hline
    Baseline & 88.55\\
    Baseline + Color Jitter & 88.63\\
    Baseline + Affine + HFlip & 94.59\\
    Baseline + Affine + Color Jitter + HFlip & 94.96\\
    \hline
    Our model(color) & 94.18\\
    Our model(color) + Color Jitter & 94.63\\
    Our model(affine + color) + HFlip & 95.16\\    
    Our model(affine + color) + HFlip + Color Jitter & \textbf{95.18}\\    
\end{tabular}%
\caption{\textbf{Impact of color and affine transformations on classification accuracy on CIFAR10.} Transformations in parentheses are learned, others are predefined. For this dataset, both color and affine transformations improve the classification accuracy. Best performances are obtained with a combination of transformations of both types.}
\label{tab:fulltransform_cifar10}
\end{table}

In a first experiment, we study color transformations alone and in combination with affine transformations on CIFAR10. For the predefined color jitter, we use the same settings as ~\cite{Cubuk_2019_CVPR}. We consider 2 versions of our model, the first one learning only color transformations and the second one learning color and affine transformations. In Tab.~\ref{tab:fulltransform_cifar10}, we can see that in both cases, the learned transformations are yielding better results than predefined ones, which illustrates the efficiency of our approach for color transformations. The best results are obtained when combining color and affine transformations.
\begin{table}[t]
\footnotesize
\centering
\begin{tabular}{l|c|c}
    ResNet18 / BACH &  Acc.\\
    \hline
    Baseline & 49.50\\
    Baseline + Color Jitter & 43.90\\    
    Baseline + VFlip & 46.00\\
    Baseline + VFlip + Color Jitter  & 44.90 \\    
    Baseline + Affine + VFlip & 50.60\\
    Baseline + Affine + VFlip + Color Jitter & 43.00 \\
    \hline    
    Our model(color) & 54.60\\
    Our model(color) + Color Jitter & 52.40\\    
    Our model(affine + color) + VFlip & \textbf{56.50}\\
    Our model(affine + color) + VFlip + Color Jitter & 49.70\\    
\end{tabular}%
\caption{\textbf{Impact of color and affine transformations on classification accuracy on BACH}. Transformations in parentheses are learned, others are predefined. Heuristically chosen color jitter parameters have a negative impact on the classifier accuracy whereas learned color transformations improve the training. Best performances are obtained with a combination of learned color and affine transformations.}
\label{tab:fulltransform_bach}
\end{table}

In a second experiment, we repeat the same protocol as in the previous experiment but on the BACH dataset. As there is no usual color jitter value for this dataset, we use the same default setting as for CIFAR10. In Tab.~\ref{tab:fulltransform_bach}, we can see that using a heuristically chosen color jitter in the predefined data augmentation leads to a significant degradation of the classifier performance. This confirms that a good data augmentation strategy in one domain is not always transferable to another and that it is safer to let the model learn the optimal transformations. This also shows that for histological images, classifiers are very sensitive to color modifications.
Best results are obtained by learning a combination of color and affine transformations.

\begin{figure}[t]
  \centering
  \includegraphics[width=\linewidth]{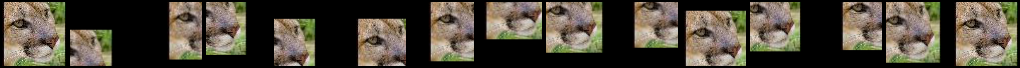}
  \includegraphics[width=\linewidth]{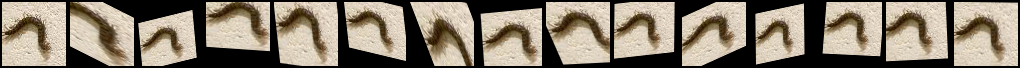}
  \includegraphics[width=\linewidth]{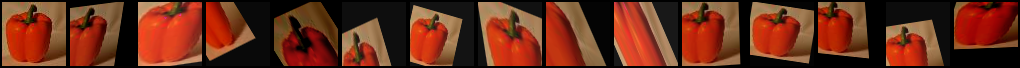}
  \includegraphics[width=\linewidth]{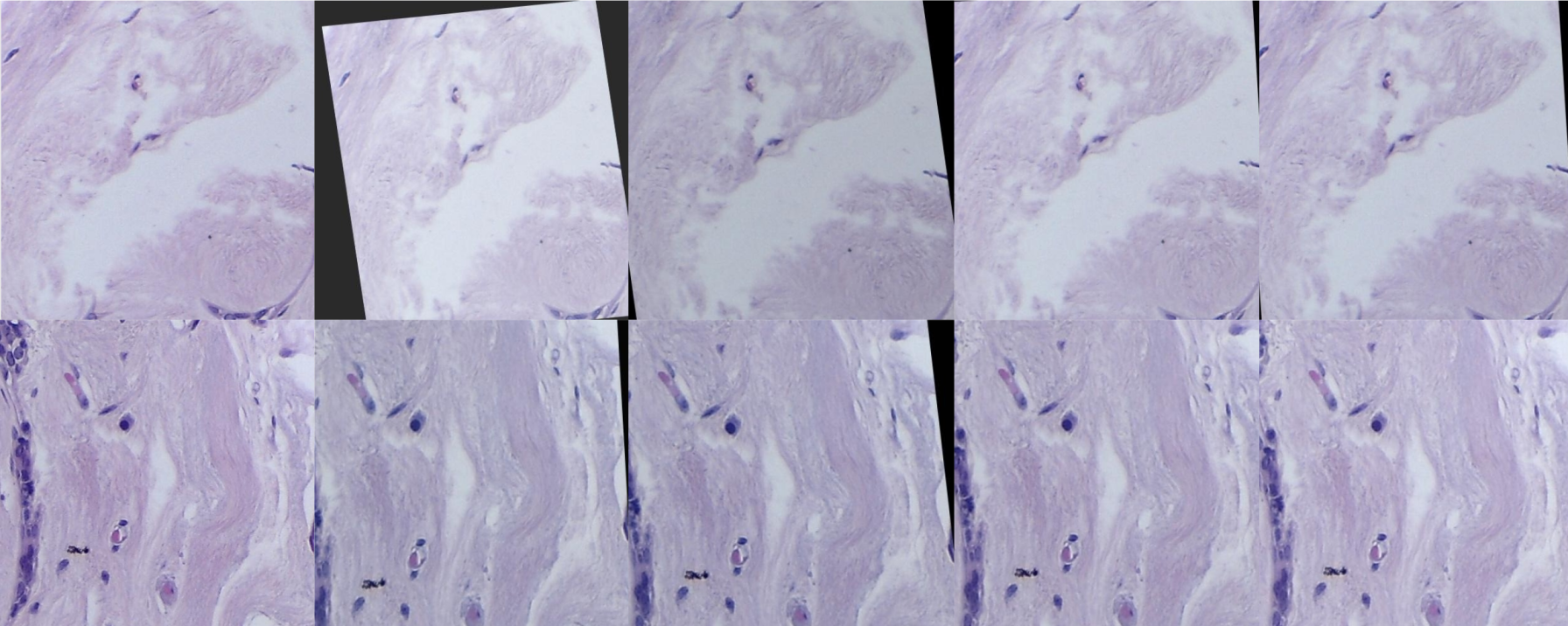}
  \includegraphics[width=\linewidth]{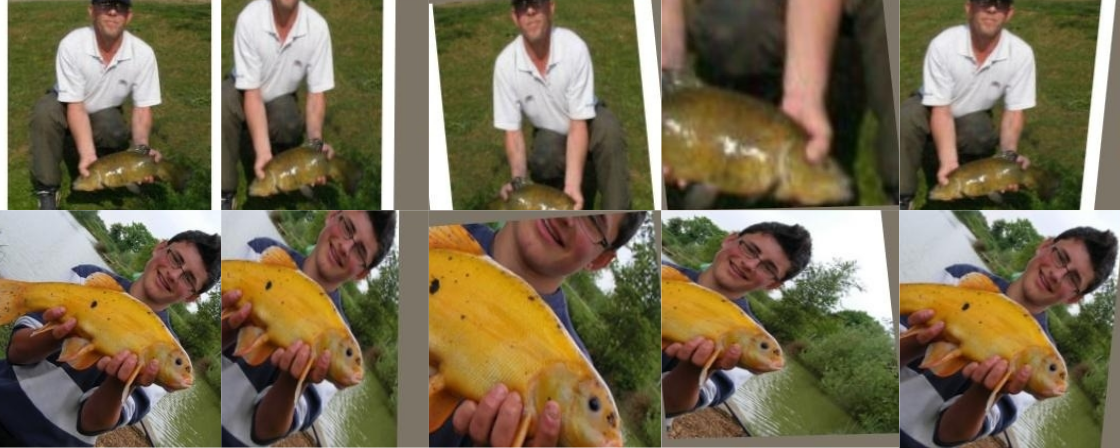}
  \caption{\textbf{Qualitative results.} The images of the first column are original images, the following ones are images transformed by our augmenter at different epochs. The first three rows contain images from Tiny ImageNet, the next two rows from BACH, and the last two rows from ImageNet.}
  \label{fig:qualitative}
\end{figure}

\vspace{-1mm}
\subsection{Evaluation on Different Datasets}
\label{subsec:eval_diffdatasets}
We now evaluate our approach on different datasets. In addition to the CIFAR10 and BACH results, we report in Tab.~\ref{tab:other_datasets} also results on CIFAR100, Tiny ImageNet and ImageNet. Predefined transformations used for CIFAR100 and Tiny Imagenet are the same as defined for CIFAR10 in \ref{subsec:augmenter_networks}. Predefined transformations for Imagenet are a resize to 256x256 followed by a random crop of size 224x224, horizontal flip and color transformations as in \cite{He2015ResNet} . Results show that our model performs better than a classifier trained only with predefined transformations on the five datasets considered already with learned \emph{Affine} transformations, but performances are even better when adding color transformations (\emph{Full}). This shows that our approach can be applied to datasets with different characteristics. Note that it is suitable not only for large scale datasets like ImageNet, but also for high resolution images like in BACH (2048x1536).

\begin{table}[t]
\centering
\resizebox{\columnwidth}{!}{%
\begin{tabular}{l|c|c|c|c|c}
                  & \textbf{CIFAR10} & \textbf{CIFAR100} & \textbf{Tiny ImageNet} & \textbf{ImageNet} & \textbf{BACH}\\
                  & \textbf{ResNet18} & \textbf{ResNet18} & \textbf{ResNet18} & \textbf{ResNet50} & \textbf{ResNet18}\\                  
    \hline
    Baseline      & 88.55 & 68.99 & 59.69 & 69.39 & 49.5\\
    Predefined    & 94.69 & 73.61 & 61.10 & 76.02 & 50.6\\\hline   
    Ours (affine) & 95.16 & 74.31 & 62.92 & 76.10 & 55,7\\
    Ours (full)   & \textbf{95.42} & \textbf{76.10} & \textbf{63.61} & \textbf{76.20} & \textbf{56.5} \\
\end{tabular}%
}
\caption{\textbf{Accuracy of our model on different datasets.} ImageNet results reported are Top1. For all datasets, our model performs better than a classifier trained only with standard predefined data augmentation.}
\label{tab:other_datasets}
\end{table}

\vspace{-1mm}
\subsection{Comparison with SotA Methods}
In Tab.~\ref{tab:comparison_sota}, we compare our model to state-of-the-art methods on CIFAR10, CIFAR100 and ImageNet. Predefined transformations are the same as described in \ref{subsec:eval_diffdatasets}. Results show that on CIFAR10 and using ResNet18 as classifier, our method obtains a better accuracy than GAN-based  automatic data augmentation learning methods. AutoAugment has a slightly better accuracy, but note that our model obtains very close results with a smaller network.
On bigger networks like Wide ResNet 28-10 and ResNet 50, our approach performs very close to search-based methods. The performance gap is explained by the fact that the search-based methods are using more transformations, in particular non-differentiable transformations, to train the end classifier. On the other end, our model requires less prior knowledge as it does not require to define a list of possible transformation and to perform an additional loop to learn the best augmentation policy from this predefined list. Considering this, it represents an interesting trade-off between training speed and accuracy, especially for datasets where potentially useful augmentations are not trivial to define.\\
On Fig.~\ref{fig:qualitative}, we show some examples of transformations learned during the training process. 
The first 3 rows show examples on Tiny ImageNet. What is interesting to note is that at the beginning of the training (left) transformations tend to be strong, while towards the end of the training (right) they are smaller and tend to approach identity. This behavior can also be seen during training on BACH (row 4 and 5) and ImageNet (row 6 and 7).

\begin{table}[t]
\centering
\resizebox{\columnwidth}{!}{%
\begin{tabular}{l|c|c|c|c}
      & \textbf{Classifier} & \textbf{CIFAR10} & \textbf{CIFAR100}& \textbf{ImageNet}\\
    \hline
    Baseline                                         & ResNet18 & 88.55 & 68.99 & -\\
    Predefined                                       & ResNet18 & 91.18 & 73.61 & - \\
    \hline
    Bayesian DA~\cite{tran2017bayesian}              & ResNet18 & 91.00 & 72.10 & -\\
    DAN~\cite{Mounsaveng2019AdversarialLO}           & BadGAN   & 93.00 & -  & -\\
    TANDA~\cite{ratner2017learning}                 & ResNet56 & 94.40 & -  & - \\
    AutoAugment~\cite{Cubuk_2019_CVPR}               & ResNet32 & \textbf{95.50} & -     & - \\\hline
    Ours                                             & ResNet18 & {95.42} & \textbf{74.31} & - \\
    \hline
    Baseline & WRN 28-10      & 94.83 & 69.90 & -\\
    Predefined & WRN 28-10    & 95.76 & 81.10 & -\\
    AutoAugment & WRN 28-10   & \textbf{97.40} & 82.90 & -\\
    Fast AA & WRN 28-10       & 97.30 & 82.70 & -\\
    PBA & WRN 28-10           & \textbf{97.40} & \textbf{83.30} & -\\
    RandAugment & WRN 28-10   & 97.30 & \textbf{83.30} & -\\\hline   
    Our model & WRN 28-10     & 96.44 & 81.90 & -\\
    \hline
    Baseline & ResNet50      & - & - & 69.39/89.41\\
    Predefined & ResNet50      & - & - & 76.02/92.84\\
    Faster AA & ResNet50       & - & - & 76.50/93.20\\    
    AutoAugment & ResNet50   & - & - & \textbf{77.60/93.80}\\
    Fast AA & ResNet50       & - & - & 77.60/93.70\\
    RandAugment & ResNet50   & - & - & \textbf{77.60/93.80}\\\hline   
    Our model & ResNet50     & - & - & 76.20/92.90\\

\end{tabular}%
}
\caption{\textbf{Comparison with other models}. ImageNet results reported are Top1/Top5. Our model based on affine and color transformations outperforms previous GAN-based models and performs at a level very close to search-based approaches. Those approaches perform better by considering also non-differentiable transformations but our approach requires less prior knowledge and no policy search loop, which makes it easier to train and more suitable for datasets where predefining data augmentation is not trivial.
}
\label{tab:comparison_sota}
\end{table}

\vspace{-2mm}
\section{Conclusion}
\label{sec:conclusion}
We have presented a novel approach to automatically learn the transformations needed for effective data augmentation. It is based on an online approximation of the bilevel optimization problem defined by alternating between optimizing the model parameters and the data augmentation hyperparameters. Thus, we can train the classifier network and an augmenter network jointly to generate the right transformations at every epoch. We evaluated the proposed approach with different models against a variety of datasets and transformations. The obtained results were comparable or better than the results obtained from defining hand-engineered transformations. This approach brings us a step closer to having a fully automated learning system that requires minimal human intervention.

\vspace{-2mm}
\section{Acknowledgment}
\label{sec:acknowledgment}
This research was supported by the National Science and Engineering Research Council of Canada (NSERC), via its Discovery Grant program and MITACS via its Acc\'el\'eration program.

{\small
\bibliographystyle{abbrvnat}
\bibliography{egbib}
}

\end{document}


\title{Learning Data Augmentation with Online Bilevel Optimization for Image Classification}

\author{Saypraseuth Mounsaveng$^{*1}$, Issam Laradji$^2$, Ismail Ben Ayed$^1$, David V\'azquez$^2$, Marco Pedersoli$^1$\\
$^1$\small{\textit{\'ETS Montreal}}
$^2$\small{\textit{Element AI}}\\
} 

\maketitle

\section{Implementation details}

We use PyTorch to implement our experiments, and Kornia library for the color transformations.

\subsection{Model architecture}
Tab.~\ref{tab:C} shows  the BadGAN classifier architecture. Tab.~\ref{tab:A} shows the augmenter network for affine and color transformations.

\begin{table}[ht]
\centering
\footnotesize
\begin{tabular}{c}
\textbf{Classifier C}\\
\hline
Input 32x32 Image\\
\hline
3x3 conv. 96 LReLU(0.2)\\
3x3 conv. 96 LReLU(0.2)\\
3x3 conv. 96 LReLU(0.2), 0.5 dropout\\
3x3 conv. 192 LReLU(0.2)\\
3x3 conv. 192 LReLU(0.2)\\
3x3 conv. 192 LReLU(0.2), 0.5 dropout\\
3x3 conv. 192 LReLU(0.2)\\
3x3 conv. 192 LReLU(0.2)\\
3x3 conv. 192 LReLU(0.2)\\
MLP 10 unit, sigmoid\\
10-class Softmax\\
\end{tabular}
\caption{\textbf{BadGAN classifier network}.}
\label{tab:C}
\end{table}

\begin{table}[ht]
\centering
\footnotesize
\begin{tabular}{c|c|c}
\multicolumn{3}{c}{\textbf{Augmenter A}}\\
\hline
\textit{Small} & \textit{Medium} & \textit{Large}\\
\hline
Input $n^*$ dim. & Input 100 dim. & Input 100 dim.\\
\hline
MLP $n$ units & MLP 64 unit & MLP 512 unit\\
relu, 0.2 dropout & relu, 0.2 dropout & relu, 0.2 dropout\\
MLP 10 x $n$ units & MLP 32 unit & MLP 1024 unit\\
relu, 0.2 dropout & relu, 0.2 dropout & relu, 0.2 dropout\\
- & - & MLP 1024 unit\\
- & - & relu, 0.2 dropout\\
- & - & MLP 512 unit\\
- & - & relu, 0.2 dropout\\
\hline
\multicolumn{3}{c}{MLP $n$ units, tanh}\\
\end{tabular}
\caption{\textbf{Augmenter network for affine and color transformations}. For \textit{Small}, $n$ is the number of parameters to learn (6 for affine, 4 for color and 10 when combining both).}
\label{tab:A}
\end{table}

\subsection{Affine transformation parameters}
In our experiments with affine transformations, the augmenter network learns a 2x3 matrix containing the parameters of the affine transformations.

\subsection{Color transformation parameters}
In our experiments with color transformations, the augmenter network outputs 1 value per transformation. Those values are in a range depending on the transformation: i) Hue in range [-0.5:0.5]; ii) Saturation in range [0:1]; iii) Contrast in range [-1:1] iv) Brightness in range [0:1]. In all cases the zero value corresponds to the identity transformation.

\section{Additional experiments}
We investigate the influence of augmenter regularization on the model performance.
Figure~\ref{fig:regularization} shows the model performance for different weight decay values. This indicates that the model is not sensitive to regularization.

\begin{figure}[h]
    \centering
    \resizebox{\columnwidth}{!}{%
    \begin{tikzpicture}
        \begin{axis}[
            legend cell align={left},
            xtick=data,
            width=\columnwidth,
            height=.4\columnwidth,
            symbolic x coords={0.001, 0.01, 0.1},
            bar width=10pt,
            enlarge x limits=0.25,
            ybar,
            ymajorgrids,
            ymin=90,
            ylabel= Accuracy (\%),
            xlabel= Weight decay value,
            legend style={at={(0.5,01.50)},
    anchor=north,legend columns=3},
        ]
        \addplot[fill=blue] coordinates {(0.1, 94.84) (0.01, 94.86) (0.001, 94.39)};
        \addplot[fill=green] coordinates {(0.1, 92.48) (0.01, 91.73) (0.001, 91.43)};
        \addplot[fill=orange] coordinates {(0.1, 94.81) (0.01, 94.93) (0.001, 94.43)};        
        \legend{affine, color, full }
        \end{axis}
    \end{tikzpicture}
    }
        \caption{\textbf{Augmenter regularization.} Accuracy for different weight decay values for the augmenter (affine, color and the combination of both) with ResNet18 on CIFAR10.}
    \label{fig:regularization}
    \vspace{-4mm}
\end{figure}


